# Super-resolution of clinical CT volumes with modified CycleGAN using micro CT volumes

Tong ZHENG[*1], Hirohisa ODA[*1], Takayasu MORIYA[*1],
Takaaki SUGINO[*1], Shota NAKAMURA[*2], Masahiro ODA[*1],
Masaki MORI[*3], Hirotsugu TAKABATAKE[*4],
Hiroshi NATORI[*5], and Kensaku MORI[*1,6,7]

**Abstract**

This paper presents a super-resolution (SR) method with unpaired training dataset of clinical CT and micro CT volumes. For obtaining very detailed information such as cancer invasion from pre-operative clinical CT volumes of lung cancer patients, SR of clinical CT volumes to μCT level is desired. While most SR methods require paired low- and high-resolution images for training, it is infeasible to obtain paired clinical CT and μCT volumes. We propose a SR approach based on CycleGAN, which could perform SR on clinical CT into μCT level. We proposed new loss functions to keep cycle consistency, while training without paired volumes. Experimental results demonstrated that our proposed method successfully performed SR of clinical CT volume of lung cancer patients into μCT level.

## 1. Introduction

Lung cancer causes largest number of deaths per year among cancers of male [1]. Currently, precise diagnosis of lung cancer mainly depends on clinical CT volumes. However, we could not obtain enough pathological information due to its low resolution. Super-resolution (SR) of clinical CT into μCT-like level is desired.

Deep learning-based methods have been proved to outperform other methods in SR. These approaches are often supervised, requiring aligned pairs of low-resolution (LR) and high-resolution (HR) patches to train a model. However, it is infeasible to obtain spatially corresponding patch pairs of clinical CT and μCT because registration between them is difficult. SR methods that can be trained by using unpaired images are desired.

---

*1 Graduate School of Informatics, Nagoya University〔Furou-cho, Chikusa-ku, Nagoya 464-0814, Japan〕
e-mail: tzheng@mori.m.is.nagoya-u.ac.jp

*2 Nagoya University Graduate School of Medicine

*3 Sapporo-Kosei General Hospital

*4 Sapporo Minami-sanjo Hospital

*5 Keiwakai Nishioka Hospital

*6 Information Technology Center, Nagoya University

*7 Research Center of Medical Bigdata, National Institute of Informatics

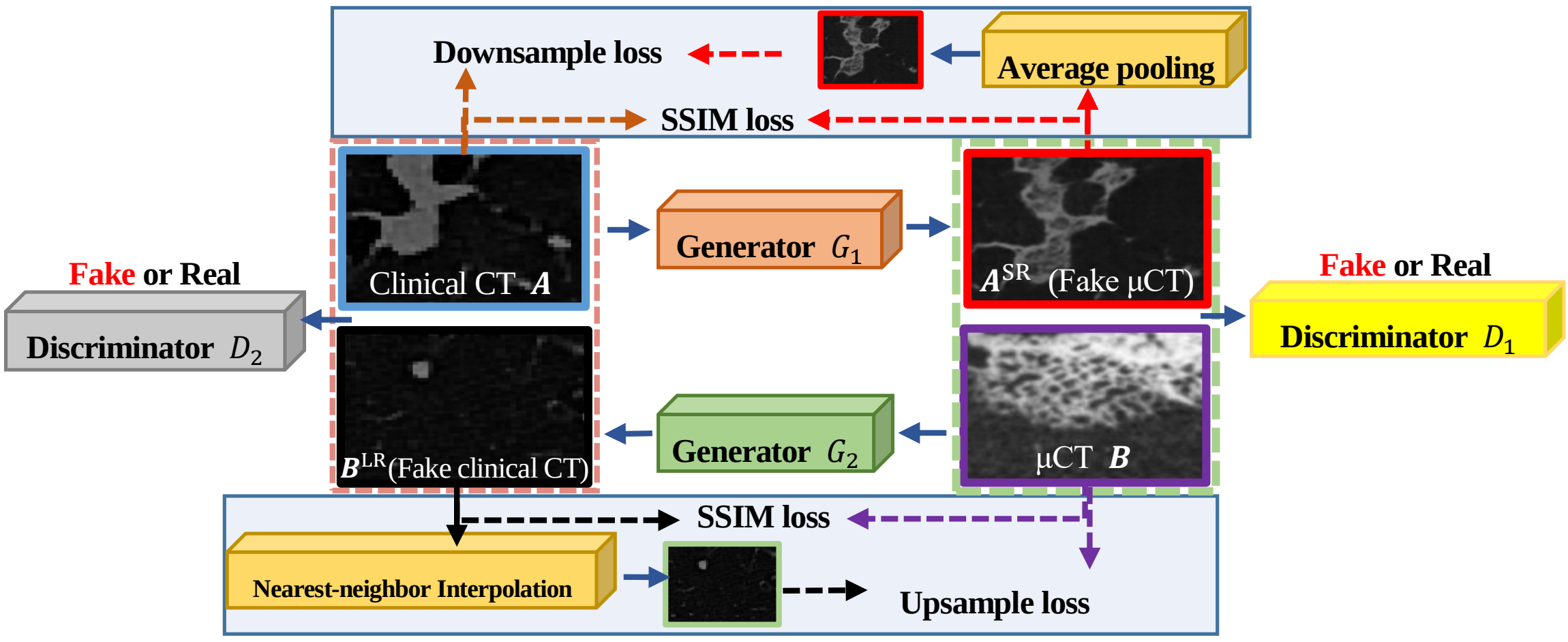

**Figure 1** Network structure used for training. Compared to original CycleGAN, proposed method uses loss functions for maintaining cycle consistency. We calculate SSIM loss between $\boldsymbol{A}$ and $\boldsymbol{A}^{\mathrm{SR}}$, $\boldsymbol{B}$ and $\boldsymbol{B}^{\mathrm{LR}}$. Further, we calculate downsample loss between average-pooled $\boldsymbol{A}^{\mathrm{SR}}$ and $\boldsymbol{A}$, as well as upsample loss between $\boldsymbol{B}$ and upsampled $\boldsymbol{B}^{\mathrm{LR}}$.

One of the first approaches that formalizes the possibility to transpose from a source domain to a target domain in the absence of paired examples is called CycleGAN [2]. For instance, pictures of the zebra are converted into those of the horse. Nevertheless, CycleGAN is not designed for SR.

In this paper, we propose an SR method of clinical CT into μCT-level by our modified CycleGAN. Unpaired clinical CT and μCT volumes are used for training.

## 2. Overview

In prior to inference, training of the network is required using clinical CT and μCT volumes. For inference, patches clipped from clinical CT volumes are input. In our study, scale of original μCT volumes is at least 8-times larger than the clinical CT volumes. Because of this, we consider 8-times SR to be the most proper.

Input of our networks are 2D patches clipped from the volumes. The input clinical CT patch size is 32×32 pixels, while input μCT patch size is 256×256 pixels.

1) Network Structure

Figure 1 shows the network structure of our proposed method. The first input is clinical CT patch $\boldsymbol{A}$, and generator $G_1$ generates corresponding SR patch $\boldsymbol{A}^{\mathrm{SR}}$ from $\boldsymbol{A}$. Similar to CycleGAN, the generator $G_1$ is aimed to produce image patches that are similar to the ones in the target domain (μCT domain) by trying to fool the discriminator $D_1$. Vice versa, the same work is done upon μCT patch $\boldsymbol{B}$ by using generator $G_2$ and discriminator $D_2$ to keep the consistency of proposed framework.

2) Loss functions

Like CycleGAN, our method uses cycle consistency while training the network. However, in SR problem, the cycle consistency between corresponding LR and SR image is different with that in image translation problem. In SR problem, corresponding LR and SR image are desired to have similarity in structure and average intensity, while the loss function used in original CycleGAN could not obtain this.

Here we propose serval loss functions in our pipeline to create the cycle consistency (blue blocks in Fig. 1). Without cycle consistency, the network would simply produce arbitrary patch in the target domain with no relationship to the structures contained in the input patch.

The first loss function we proposed to keep cycle consistency is downsample loss. It is defined to maintain similarity while transforming clinical CT volume to μCT scale as

$$l_{\mathrm{downsample}}(\boldsymbol{A}) = \mathrm{MSE}(\boldsymbol{A}, f(\boldsymbol{A}^{\mathrm{SR}})),$$

where $f()$ is an average pooling function, reducing the size of $\boldsymbol{A}^{\mathrm{SR}}$ to the same as $\boldsymbol{A}$, since $\boldsymbol{A}^{\mathrm{SR}}$ is SR patch, 8 times larger than $\boldsymbol{A}$. MSE is the mean squared error. Analogously, we name the second loss function the upsample loss as

$$l_{\mathrm{upsample}}(\boldsymbol{B}) = \mathrm{MSE}\big(\boldsymbol{B}, g(\boldsymbol{B}^{\mathrm{LR}})\big),$$

where $g()$ is the nearest-neighbor interpolation function, upsampling the size of generated clinical-CT like $\boldsymbol{B}^{LR}$ to the original size of $\boldsymbol{B}$.

Although the first and second loss function could keep the cycle consistency while training network, both loss functions depend on intensity differences between generated and target image patches, which is not very well matched to perceived visual quality. Here we propose third and fourth loss functions, which we name as clinical-SSIM loss and micro-SSIM loss

$$l_{\mathrm{clinical-SSIM}} = \frac{1}{\mathrm{SSIM}(\boldsymbol{A}, f(\boldsymbol{A}^{\mathrm{SR}}))},$$

$$l_{\mathrm{micro-ssim}} = \frac{1}{\mathrm{SSIM}(\boldsymbol{B}, g(\boldsymbol{B}^{\mathrm{LR}}))},$$

where SSIM is the structural similarity proposed in paper [3]. While training our model, third and fourth loss function helps protecting the model from generating blurred image patches.

3) Training

We perform training process using 2D clinical CT patches as input of Generator $G_1$ and 2D μCT patches as input of Generator $G_2$. Output of Generator $G_1$ is the generated μCT-like SR patches. Discriminator $D_1$ is used to discriminate output of Generator $G_1$ is real or fake. Furthermore, for more stable training, we mixed downsampled μCT patches in clinical CT patches as input. The percentage of downsampled μCT patches is 25%.

4) Inference

For testing, we input 2D patch clipped from clinical CT volumes into the trained Generator $G_1$. Output is a SR patch based on the input patch.

**Table 1** Profiles of clinical CT and μCT volumes

| | μCT | Clinical CT |
|---|---|---|
| pixels in one slice | 1024×1024 pixels | 512×512 pixels |
| number of slices | 545~1082 slices | 435~554 slices |
| size of each pixel | 34~53μm | 0.625mm |
| slide thickness | 34~53μm | 0.6mm |

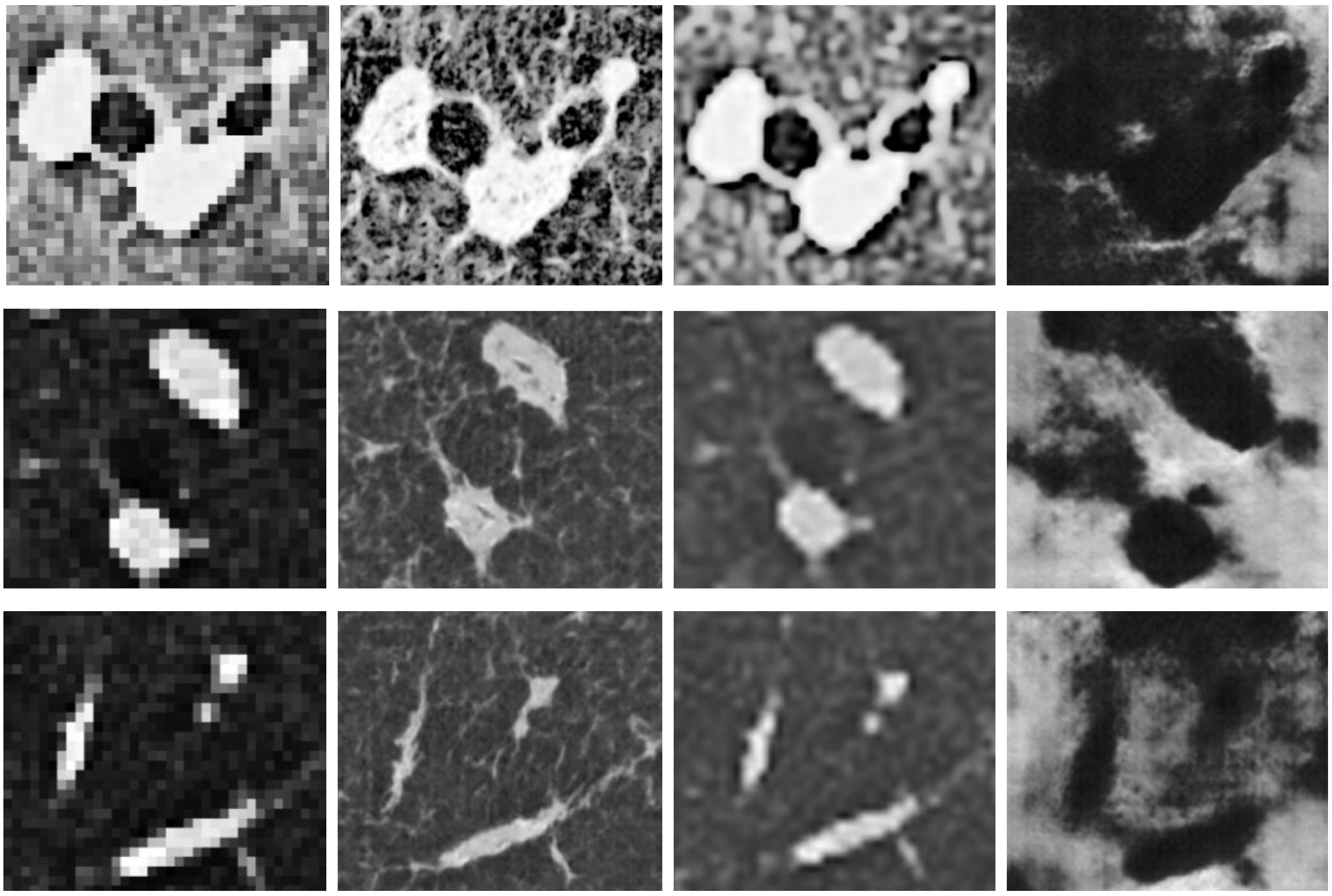

**Figure 2** Example and comparison of proposed method. (a) original clinical CT, (b) proposed SR result, (c) bicubic interpolation, and (d) CycleGAN (without proposed loss functions).

## 3.Experiments and results

We utilized five lung cancer cases in the experiment. Clinical CT volumes were acquired by SOMATOM Definition flash (SIEMENS, Germany) from lung cancer patients. After surgical dissection of the lung cancers, μCT volumes of the dissected specimens were acquired by a μCT scanner, inspeXio SMX-90CT Plus (SHIMADZU, Japan). The profiles of clinical CT and μCT volumes are listed in Table 1.

In our experiment, we used clinical CT and μCT volumes obtained from same patients, meaning one patient has one clinical CT volume and one μCT volume. We used 5 cases of clinical CT and μCT for training in our experiment. The epoch number is 200. For testing, we used 1case of clinical CT.

SR results of our proposed method were compared to bicubic-interpolation and original CycleGAN, as shown in Fig. 2. We could obtain more details from SR results than bicubic-interpolation results. Lung anatomies, such as the bronchus looks more clearly than bicubic-interpolation. Original CycleGAN's result has produced very different results from original clinical CT volumes.

## 4. Discussion and conclusion

We proposed a novel SR method with unpaired training dataset of clinical CT and micro CT volumes. New loss functions are introduced to keep cycle consistency in SR task. Experimental result showed that our method could apply

SR on clinical CT to μCT level.

Because training of proposed method is unpaired, we do not have corresponding ground truth for certain input, quantitative evaluation of output result becomes difficult. Our future work is quantitative evaluation of SR results.

**Competing interests**

None.

**Acknowledgement**

Parts of this research was supported by MEXT·JSPS KAKENHI (26108006, 17H00867, 17K20099), the JSPS Bilateral International Collaboration Grants, the AMED18lk1010028s0401, the AMED19lk1010036h0001 and the Hori Sciences & Arts Foundation.

# μCT を用いた改良版 Cycle-GAN による臨床用 CT 像の超解像処理

鄭 通[*1], 小田 紘久[*1], 守谷 享泰[*1], 杉野貴明[*1], 中村 彰太[*2], 小田 昌宏[*1],

森 雅樹[*3], 高畠博嗣[*4], 名取 博[*5], 森 健策[*1,6,7]

*1 名古屋大学大学院情報学研究科

*2 名古屋大学大学院医学系研究科

*3 札幌厚生病院

*4 札幌南三条病院

*5 恵和会西岡病院

*6 名古屋大学情報基盤センター

*7 国立情報学研究所医療ビッグデータ研究センター

本稿では，臨床用 CT 像の超解像手法を提案する．肺がん症例の臨床 CT 像から腫瘍の浸潤状況など疾患に関する情報を取得するため，肺臨床 CT に超解像を適用し，μCT レベルの解像度を得る手法が求められている．多くの超解像手法における教師あり学習では，対応関係のある低解像度と高解像度の画像ペアが必要となるが，臨床 CT と μCT の画像ペアは正確に位置合わせを行うことが困難である．我々は CycleGAN を改良し，臨床用 CT 像と μCT 像間の相互変換において一貫性を保持するための新しい損失関数を導入することによって，ペアなしの超解像手法を実現する．実験の結果，臨床 CT 像の μCT レベルへの超解像が可能であった．